\documentclass[11pt]{article}

\usepackage[preprint]{acl}

\usepackage{times}
\usepackage{latexsym}
\usepackage{amssymb}
\usepackage{amsmath}
\usepackage{geometry}
\usepackage{enumitem} 
\usepackage{everysel} 
\usepackage{setspace}
\usepackage{multicol}
\usepackage{algorithm}
\usepackage[noend]{algpseudocode}
\usepackage{amsmath}
\usepackage{xcolor}
\usepackage{graphicx}
\usepackage{framed}
\usepackage{inconsolata}
\usepackage{booktabs}
\usepackage{multirow}
\usepackage{makecell}
\usepackage{caption}

\usepackage[T1]{fontenc}
\usepackage[utf8]{inputenc}

\usepackage{microtype}

\usepackage{inconsolata}

\usepackage{graphicx}

\usepackage[table]{xcolor}
\usepackage{booktabs}
\usepackage{xcolor}
\usepackage{transparent}
\definecolor{nihongblue}{HTML}{4D4DFF}
\definecolor{nihongpink}{HTML}{FF6EC7}
\definecolor{deepbrown}{HTML}{97694F}
\definecolor{forestgreen}{HTML}{238E23}
\definecolor{bluepurple}{HTML}{9F5F9F}
\definecolor{jiaohong}{HTML}{cf292f}
\definecolor{conglv}{HTML}{94c66b}
\definecolor{SR}{HTML}{D9988B}
\definecolor{RR}{HTML}{9BB7D4}
\definecolor{NR}{HTML}{A9C1AA}
\title{"You Are Rejected!": An Empirical Study of Large Language Models \\ Taking Hiring Evaluations}

\author{
  Dingjie Fu\textsuperscript{1}$^{*}$, Dianxing Shi\textsuperscript{2}$^{*}$
  \\ \textsuperscript{1}Huazhong Univeristy of Science and Technology (HUST), \textsuperscript{2}Independent Researcher
  \\ 
  \small{$^{*}$ Equal Contribution} \quad
  \small{
   \textbf{Correspondence:} 
   \href{dingjiefu1103@gmail.com }{dingjiefu1103@gmail.com}
 }
}

\begin{document}
\maketitle
\begin{abstract}
With the proliferation of the internet and the rapid advancement of Artificial Intelligence, leading technology companies face an urgent annual demand for a considerable number of software and algorithm engineers. To efficiently and effectively identify high-potential candidates from thousands of applicants, these firms have established a multi-stage selection process, which crucially includes a standardized hiring evaluation designed to assess job-specific competencies. Motivated by the demonstrated prowess of Large Language Models (LLMs) in coding and reasoning tasks, this paper investigates a critical question: \textbf{\textit{Can LLMs successfully pass these hiring evaluations?}} To this end, we conduct a comprehensive examination of a widely-used professional assessment questionnaire. We employ state-of-the-art LLMs to generate responses and subsequently evaluate their performance. Contrary to any prior expectation of LLMs being ideal engineers, our analysis reveals a significant inconsistency between the model-generated answers and the company-referenced solutions. Our empirical findings lead to a striking conclusion: \textbf{\textit{All evaluated LLMs fails to pass the hiring evaluation.}} 
\end{abstract}

\section{Introduction}
\begin{figure*}[ht]
  \centering
   \includegraphics[width=1.0\linewidth]{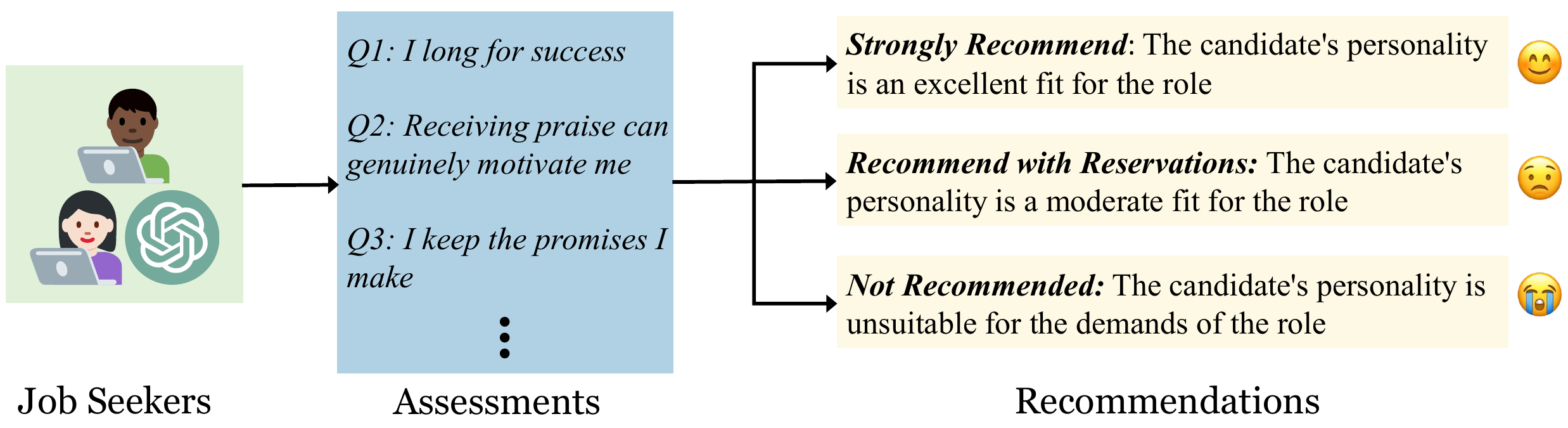}
   \caption{The hiring evaluation process involves two stages: (1) candidates responding to a standardized questionnaire, (2) the system provides a recommendation based on their answers.}
   \label{fig:illustration}
\end{figure*}

The past decade has witnessed the expansion of the internet and the remarkable progress in Artificial Intelligence (AI), fueling a high demand for Software Develop Engineers (SDEs). Consequently, technology companies open thousands of positions to secure talented individuals \citep{raghavan2020mitigating, an-etal-2024-large}. However, for a single job opening, the hiring sector may need to review applications from tens of hundreds of candidates. To weed out unqualified applicants, the hiring evaluation, a standardized assessment within a thorough selection process, is widely employed by organizations to profile candidates, as illustrated in Figure \ref{fig:illustration}. 

Recent advances in Large Language Models (LLMs) have led to significant improvements beyond Natural Language Processing (NLP) \citep{guo2025deepseek, yuan2025native}. State-of-the-art models such as GPT-5 and Gemini-2.5 demonstrate a remarkable capability for tackling diverse and complicated tasks. In domains like academic writing \citep{wang2025llms} and code generation \citep{qwen3technicalreport}, these models have served as powerful assistants. Given that LLMs increasingly exhibit competencies of a talented SDE---including proficiency in problem decomposition, reasoning, and programming---this trend motivates a critical question: \textit{can LLMs pass the hiring evaluations established by technology companies?} This research focuses on hiring evaluations as they represent a relatively unexplored domain compared to coding assessments, which have demonstrated high solvability by LLMs \cite{gaebler2024auditing, wu2024survey, zhao2024recommender}.

To address this question, we conduct a comprehensive evaluation using a standard professional assessment questionnaire. A typical hiring evaluation process consists of two stages: an assessment phase, where candidates respond to behavioral and preference questions, and a recommendation phase, where these responses are synthesized into a hiring decision. Our study subjects LLMs to this process to determine their feasibility as candidates. Based upon the above analysis, the goal of this paper is to answer the following research questions:

\begin{description} 
    \item[RQ1] Do LLMs possess the capability to pass a standardized hiring evaluation?
    \item[RQ2] Do different LLMs exhibit distinct personality profiles in their responses?
    \item[RQ3] Do LLMs demonstrate the competency to generate viable hiring recommendations?
\end{description}

In our study, we find that LLMs tend to produce highly idealistic answers, with certain responses standing in sharp contrast to company preferences. This divergence may indicate an inconsistency between the learned preferences of LLMs and the complexities of practical scenarios.

\section{Related Work}
While LLMs are increasingly deployed in automated hiring systems, they are also susceptible to inheriting biases present in their training data \citep{NEURIPS2020_92650b2e, blodgett-etal-2020-language, salinas2023unequal}. Consequently, substantial research efforts have been devoted to exploring hiring biases in LLMs. \citet{wang2024jobfair} presents a novel framework for benchmarking hierarchical gender hiring bias in LLMs, revealing significant issues of reverse gender hiring bias. \citet{kamruzzaman-kim-2025-impact} examines implicit age-related name bias in LLMs by showing that a candidate's name significantly affects the resulting recommendations.

Departing from this employer-centric focus, our research investigates the use of LLMs from the perspective of job-seekers. While \citet{bhattacharya2025let} introduces a multi-agent AI system to provide more trustworthy and fair hiring decisions, we employ LLMs to simulate hiring evaluations. In this paper, we conduct an in-depth analysis to identify and analyze discrepancies between the model-generated responses and the company-referenced choices.

\section{Experimental Setup}
\label{sec:setup}
\noindent\textbf{Problem Definition} \quad
The hiring evaluation process typically involves a standardized questionnaire, and the final recommendation is based on the candidate's answers. Formally, given a questionnaire $\mathcal{Q}=\left\{q_1, q_2, ..., q_n\right\}$, where $q$ refers to a question and $n$ indicates the number of questions. $\mathcal{Q}$ takes the form of brief status descriptions, and $\mathcal{A}=\left\{a_1, a_2, ..., a_n\right\}$ comprises its corresponding answers, where $a_i$ refers to the answer of $q_i$. Each $a$ is assigned a numerical score on an integer scale from $1$ to $9$, representing the degree of agreement, where higher values indicate stronger agreement. The complete questionnaire and its corresponding reference answers are provided in Appendix \ref{sec:questionnaire}.

\noindent\textbf{Data Collection} \quad
Our curated data is derived from two parts: responses generated by LLMs to specific prompts, and answers provided by $2$ volunteers. This data collection approach enables a comparative analysis of model outputs against established human norms. The detailed prompt templates and collection processes are outlined in the Appendix \ref{subsec:data_generation}. 

\noindent\textbf{Models} \quad
We deploy \textbf{12} models in our experiments namely {\transparent{0.8}\textcolor{nihongblue}{GPT-4o}}, {\transparent{1.0}\textcolor{nihongblue}{GPT-5-chat}}, {\transparent{0.6}\textcolor{nihongpink}{DeepSeek-R1}}, {\transparent{0.8}\textcolor{nihongpink}{DeepSeek3.1-Terminus}}, {\transparent{1.0}\textcolor{nihongpink}{DeepSeek3.2-Exp}}, {\transparent{0.6}\textcolor{deepbrown}{Qwen1.5-32B-chat}}, {\transparent{0.8}\textcolor{deepbrown}{Qwen2.5-32B}}, {\transparent{1.0}\textcolor{deepbrown}{Qwen3-32B}}, {\transparent{0.8}\textcolor{forestgreen}{Gemini2.5-flash}}, {\transparent{1.0}\textcolor{forestgreen}{Gemini2.5-pro}}, {\transparent{0.8}\textcolor{bluepurple}{Llama3.3-70B}}, {\transparent{1.0}\textcolor{bluepurple}{Llama4-Maverick-17B-128E-FP8}}. See Appendix \ref{subsec:llms} for more information about models.

\noindent\textbf{Evaluation Protocol} \quad
The hiring evaluation, also known as a personality test, features questions without standard solutions. To address this, model performance is assessed utilizing a set of reference answers $\mathcal{R}=\left\{r_1, r_2, ..., r_n\right\}$. Specifically, Our evaluation protocol assesses model performance through three complementary approaches: (1) a global difference measure using Root Mean Squared Error ($\text{RMSE}(\mathcal{A}, \mathcal{R}) = \sqrt{\frac{1}{N} \sum_{i=1}^{N} (a_i - r_i)^2}$), (2) a pattern similarity analysis via the Pearson Correlation Coefficient ($\gamma(\mathcal{A}, \mathcal{R}) = \frac{ \sum (a_i - \bar{a})(r_i - \bar{r}) }{ \sqrt{\sum (a_i - \bar{a})^2}  \sqrt{\sum (r_i - \bar{r})^2} }$), and (3) a fine-grained diagnosis conducted through human-participant analysis. See Appendix \ref{subsec:metrics} for detailed introduction of these metrics.

\section{Results and Discussion}
LLMs are attributed with excellent reasoning and programming capabilities, naturally deemed qualified for software engineering roles. In this study, we evaluate this claim and moving beyond perceived competencies to identify their practical limitations. Despite their demonstrated intellectual prowess, a misalignment exists between the behavioral profiles of these LLMs and the specific traits sought by companies.

\begin{table}[tbp]
  \centering
  \vspace{-1mm}
  \small
  \begin{tabular}{l c c}
    \hline
    \textbf{Model} & \textbf{RMSE($\downarrow$)} & \textbf{$\gamma$($\uparrow$)}  \\
    \hline 
    {\transparent{0.8}\textcolor{nihongblue}{GPT-4o}} & 2.4879 & 0.3830 \\
    {\transparent{1.0}\textcolor{nihongblue}{GPT-5-chat}} & 2.7797 & 0.3610 \\
    {\transparent{0.6}\textcolor{nihongpink}{DeepSeek-R1}}$^{\dagger}$ & \textbf{1.5870} & \textbf{0.7809} \\
    {\transparent{0.8}\textcolor{nihongpink}{DeepSeek3.1-Terminus}} & 2.5313 & 0.4483 \\{\transparent{1.0}\textcolor{nihongpink}{DeepSeek3.2-Exp}} & 2.3244 & 0.4944 \\
    {\transparent{0.6}\textcolor{deepbrown}{Qwen1.5-32B-chat}} & 2.3629 & 0.3770 \\{\transparent{0.8}\textcolor{deepbrown}{Qwen2.5-32B}} & 2.1344 & 0.4681 \\{\transparent{1.0}\textcolor{deepbrown}{Qwen3-32B}} & 2.6658 & 0.1599 \\{\transparent{0.8}\textcolor{forestgreen}{Gemini2.5-flash}} & 2.1060 & 0.4041 \\
    {\transparent{1.0}\textcolor{forestgreen}{Gemini2.5-pro}}$^{\dagger}$ & 2.2953 & 0.4776 \\
    {\transparent{0.8}\textcolor{bluepurple}{Llama3.3-70B}} & 2.3144 & 0.4751 \\{\transparent{1.0}\textcolor{bluepurple}{Llama4-Maverick-17B-128E-FP8}} & 2.2288 & 0.5257 \\
    \hline
  \end{tabular}
    \caption{Quantitative evaluation of various LLMs. We report the Root Mean Squared Error (RMSE) and Pearson Correlation Coefficient ($\gamma$) for 12 models, evaluated by their consistency with human references. Best results are in \textbf{boldface}, $\dagger$ denotes reasoning model.}
      \label{tab:metric}
       \vspace{-1mm}
\end{table}

\subsection{Quantitative Analysis (RQ1)}
\newcommand{\highlight}[1]{
  \par\noindent
  \fboxsep=6pt
  \colorbox{blue!5}{\parbox{\dimexpr\linewidth-2\fboxsep}{#1}}
  \par
}

\highlight{
    \paragraph{Finding 1:} 
    \emph{LLMs often produce responses that diverge from human preferences.}
}
\vspace{10pt}
We conduct a quantitative evaluation of various state-of-the-art (SOTA) LLMs. As summarized in Table \ref{tab:metric}, \textbf{\textit{all models exhibit a statistically significant discrepancy from human preferences.}} Specifically, the RMSE of non-reasoning models exceeds $2.0$ on a $1$-$9$ Likert scale, reflecting a considerable misalignment between LLM-generated responses and human expectations. Among these models, {\transparent{0.8}\textcolor{forestgreen}{Gemini2.5-flash}} achieves the lowest RMSE of $2.1060$. For reasoning models, we evaluate {\transparent{0.6}\textcolor{nihongpink}{DeepSeek-R1}} and {\transparent{1.0}\textcolor{forestgreen}{Gemini2.5-pro}}. While {\transparent{1.0}\textcolor{forestgreen}{Gemini2.5-pro}} performs similarly to {\transparent{0.8}\textcolor{forestgreen}{Gemini2.5-flash}}, we observe that {\transparent{0.6}\textcolor{nihongpink}{DeepSeek-R1}} achieves a markedly lower RMSE of $1.5870$, demonstrating its stronger comprehension of the questionnaire.

\begin{figure}[ht]
  \centering
   \includegraphics[width=1.0\linewidth]{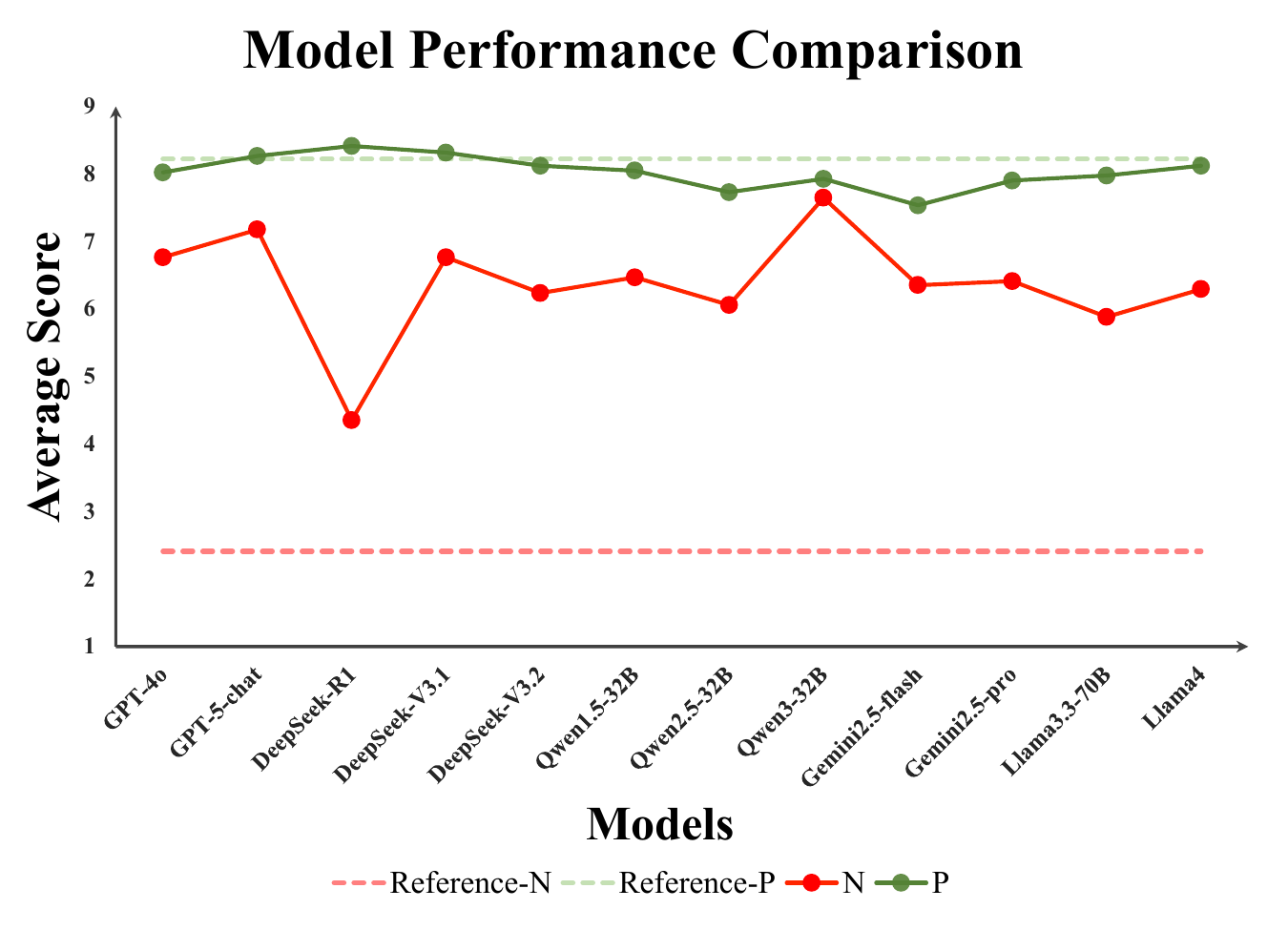}
   \caption{Average scores of LLMs versus human references on \textcolor{conglv}{positive} (P) and \textcolor{jiaohong}{negative} (N) questions. Abbreviated model names are used for clarity.}
   \label{fig:comparison}
\end{figure}

In terms of the Pearson correlation coefficient ($\gamma$), although all values are positive, only one model surpasses $0.7$. The majority of models attain $\gamma$ values between $0.3$ and $0.5$, indicating only a weak to moderate correlation with human reference answers. The highest performance is achieved by {\transparent{0.6}\textcolor{nihongpink}{DeepSeek-R1}}, showcasing its responses are highly consistent with the references. In contrast,{\transparent{1.0}\textcolor{deepbrown}{Qwen3-32B}} presents a notable exception , with a $\gamma$ of only $0.1599$, pointing to a weaker alignment.

The hiring evaluation questions can be categorized into three groups: (1) positive questions, which expect answers in a positive direction; (2) negative questions, which aim at receiving disagreement; (3) self-descriptive questions, which allow candidates to respond based on their self-perception. As shown in Figure \ref{fig:comparison}, we compare the average scores of LLMs and human references on positive and negative questions. While {\transparent{0.6}\textcolor{nihongpink}{DeepSeek-R1}} demonstrates an ability to disagree with negative questions, with an average score below $5$ (Neutral), other models fail to make this distinction apparent. The results indicate that LLMs exhibit a significant tendency to select positive options, highlighting a divergence from human preferences. As illustrated in the Appendix \ref{subsec:rq1_supple}, our findings remain robust even when the option scales are reversed (with $1$ for agreement and $9$ for disagreement).

\subsection{The Profile of LLM (RQ2)}
\highlight{
    \paragraph{Finding 2:} 
    \emph{LLMs present almost similar personality profiles.}
}
\vspace{10pt}

\begin{figure}[ht]
  \centering
   \includegraphics[width=1.0\linewidth]{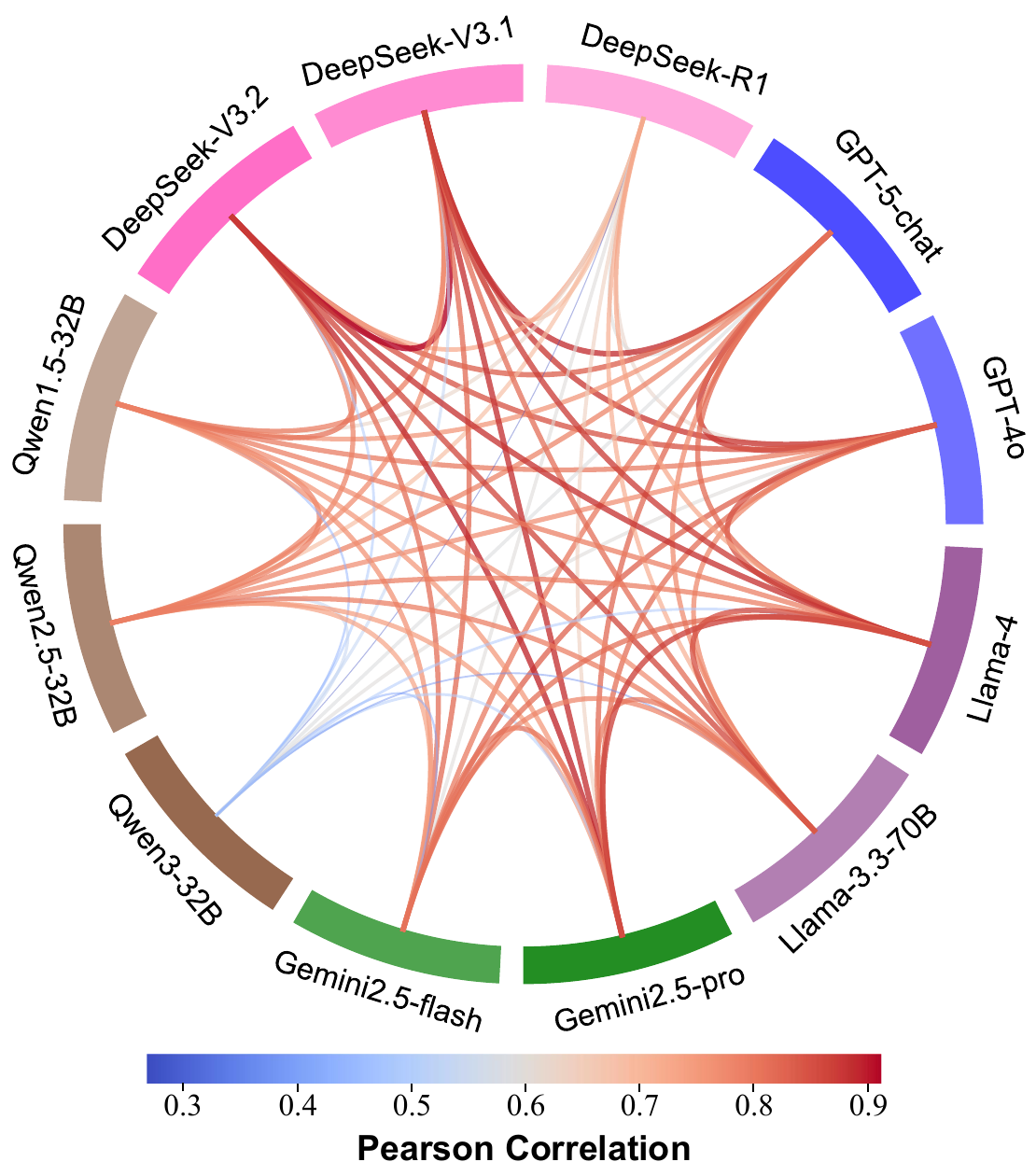}
   \caption{Pearson correlation coefficients across 12 LLMs. While LLMs exhibit highly consistent, the \textcolor{deepbrown}{Qwen3-32B} emerges as a notable outlier. Abbreviated model names are used for clarity.}
   \label{fig:pearson}
\end{figure}

To investigate the presence of distinct personality profiles in LLMs, we compute the Pearson correlation coefficients ($\hat{\gamma}$) across 12 models. As shown in Figure \ref{fig:pearson},  \textbf{\textit{the responses produced by different LLMs demonstrate striking consistency, indicating shared preferences and personality profiles.}} This high degree of alignment is evidenced by correlation coefficients predominantly exceeding $0.7$, despite substantial variations in model architectures training data, and scale properties. Such convergence suggests that current LLMs may capture similar underlying patterns from their internet-scale training corpora.

Only one notable exception: \textcolor{deepbrown}{Qwen3-32B} exhibits a pronounced divergence from the others. Its outlier status is evidenced by a low $\gamma$ value of $0.1599$ (see Table \ref{tab:metric}) and its propensity to give highly positive responses on negative questions (see Figure \ref{fig:comparison}). This result indicates a systematic misalignment in its internal representation of the questions, reflecting a substantial deviation from both practical scenarios and the understanding of other models. The complete correlation matrix and metrics are provided in Appendix \ref{subsec:rq2_supple}.

\subsection{Qualitative Analysis (RQ3)}
\highlight{
    \paragraph{Finding 3:} 
    \emph{LLMs can not generate reasonable hiring recommendations.}
}
\vspace{10pt}
To explore the feasibility of LLMs as hire resources manager, we deploy \textbf{9} models to simulate the role. Specifically, we utilize an HR prompt (see Figure \ref{fig:prompt2}) to activate HR mode and input a completed questionnaire. We then collect and analyze the hiring recommendations generated by each LLM (see Figure \ref{fig:hr-candidate} and Appendix \ref{subsec:rq3_supple}). 

As depicted in Figure \ref{fig:hr-candidate}, \textbf{\textit{the competency of LLMs in the role of HR managers appears questionable.}} We employ LLMs to generate pros and cons, final recommendations, and comprehensive reviews. According to the recommendation results, no candidate is rated as \textit{Not Recommended}. Furthermore, it is worth noting that the reference candidate receives a result of \textit{Recommended with Reservations} from 7 out of the 9 models. 

Additionally, both {\transparent{0.8}\textcolor{nihongblue}{GPT-4o}} and \textcolor{nihongblue}{GPT-5-chat} assign a \textit{Strongly Recommend} rating. However, the robustness of this positive assessment is undermined by the fact that these two models issue the identical results for all candidates, suggesting a potential lack of discriminatory judgment. We manually analyze the conclusions produced by the HR models, revealing that there exists a discrepancy with the reference answers, particularly concerning low-scoring responses. This finding is consistent with the results in Figure \ref{fig:comparison}, which illustrates a tendency for LLMs to converge on "Agreeable" responses for all questions.

These outcomes stem from the LLMs' limited capacity to comprehend the intricate relationship between the positive/negative valences of the questions and the nuanced requirements of the career role. Consequently, this limitation impairs their ability to generate empirically reasonable and well-justified hiring recommendations. For more details, refer to Appendix \ref{subsec:rq3_supple}.

\section{Conclusion}
This paper presents a comprehensive examination of a professional assessment questionnaire to evaluate the suitability of LLMs as both job seekers and HR managers. Our empirical findings reveal a significant discrepancy between LLM-generated responses and human preferences. This bias demonstrates that while LLMs are trained on web-scale data, they lack the discernment to identify subtle yet critical factors. Specifically, LLMs tend to select traits associated with an "ideal worker" from a generic perspective, while neglecting the strategic needs of a company, which seeks to derive benefits from its employees. Therefore, equipping LLMs with a genuine understanding of this dual-sided reality remains a long-term challenge.

\section*{Limitations}
Our study focuses on hiring evaluation, specifically using a personality test for software development engineers. However, this work is limited by its evaluation of a single questionnaire and a specific set of LLMs, which hinders a comprehensive exploration of the area. Furthermore, our methodology did not involve multiple calls to the same model to eliminate the potential variance introduced by the temperature parameter. Future work should aim to overcome these limitations.

% \section*{Acknowledgments}

\bibliography{custom}

\clearpage
\appendix
\begin{center}
{\Large \textbf{Appendix}}
\end{center}

\section{List of Questionnaire and Reference Answers}
\label{sec:questionnaire}
We present the utilized questionnaire and its corresponding reference answers in the following. Questions highlighted in \textcolor{jiaohong}{RED} indicate negative items, while those in \textcolor{conglv}{GREEN} represent positive items.

\vspace{5pt}
\noindent\rule{\linewidth}{1pt}
\vspace{-4ex}
\newlist{quiz}{enumerate}{1}
\setlist[quiz,1]{
    leftmargin=*,
    label=\arabic*.,
    itemsep=0.5ex,
    format=\makebox[2.5em][r]{}\hspace{0.5em}
}

\footnotesize
% \begin{multicols}{2}
\begin{quiz}
    \setlength{\itemsep}{0.5pt}
    \item I long for success, \textbf{3}
    \item Receiving praise can genuinely motivate me, \textbf{5}
    \item {\color{jiaohong}I am very interested in the motivations behind people’s behavior}, \textbf{2}
    \item I feel it’s important to grasp all relevant information, \textbf{7}
    \item {\color{conglv}I feel it’s important to complete tasks before the final deadline}, \textbf{9}
    \item I hope to receive feedback from others on my performance, \textbf{6}
    \item {\color{conglv}I must understand the underlying principles to learn more effectively}, \textbf{7}
    \item {\color{conglv}I keep the promises I make}, \textbf{9}
    \item I can solve problems, \textbf{8}
    \item I am in control of my own future, \textbf{3}
    \item I am good at making friends, \textbf{6}
    \item I am good at encouraging others, \textbf{6}
    \item I gain genuine satisfaction from discovering business opportunities, \textbf{7}
    \item I like to meet strangers, \textbf{6}
    \item I feel I must consider the viewpoints of others, \textbf{6}
    \item {\color{conglv}I must find a way to solve problems}, \textbf{8}
    \item {\color{conglv}Punctuality is an important principle for me}, \textbf{9}
    \item I gain genuine satisfaction from developing strategies, \textbf{6}
    \item I can come up with many ideas, \textbf{7}
    \item {\color{conglv}I am good at following procedures}, \textbf{8}
    \item I have learned a lot through reading, \textbf{6}
    \item I am a tolerant person, \textbf{6}
    \item I can quickly build rapport with others, \textbf{7}
    \item I am good at gett ing started on work, \textbf{8}
    \item I need to be the center of attention, \textbf{3}
    \item {\color{conglv}I like to work under pressure}, \textbf{8}
    \item {\color{conglv}I tend to learn by doing}, \textbf{9}
    \item I have a strong sense of curiosity, \textbf{4}
    \item I think it’s important to make plans for things, \textbf{6}
    \item I like to propose many ideas, \textbf{7}
    \item I am good at handling uncertainty, \textbf{7}
    \item I am good at discovering how to improve things, \textbf{8}
    \item {\color{jiaohong}I am good at understanding the motivations behind people’s behavior}, \textbf{2}
    \item People say I make a good first impression, \textbf{6}
    \item I am good at negotiating, \textbf{4}
    \item {\color{jiaohong}I am ready to make important decisions at any time}, \textbf{3}
    \item I enjoy the feeling of being full of energy, \textbf{7}
    \item I want to maximize benefits as much as possible, \textbf{3}
    \item {\color{conglv}I enjoy working in a team}, \textbf{8}
    \item I believe having common sense is very important, \textbf{6}
    \item I hope things can be completed properly, \textbf{5}
    \item I am more willing to face things with a happy attitude, \textbf{7}
    \item I am willing to adapt to new challenges, \textbf{7}
    \item {\color{conglv}I still handle things well when work is busy}, \textbf{8}
    \item {\color{conglv}I have good written communication skills}, \textbf{8}
    \item I am a person who is willing to be considerate of others, \textbf{6}
    \item I am good at explaining problems clearly, \textbf{6}
    \item {\color{conglv}I will persevere even when faced with difficult challenges}, \textbf{9}
    \item I am eager to close a deal, \textbf{3}
    \item {\color{jiaohong}I really like to argue with people}, \textbf{2}
    \item {\color{conglv}I rarely feel anxious during major events}, \textbf{9}
    \item I tend to make decisions based on objective facts, \textbf{7}
    \item {\color{conglv}I need to have a system to follow when doing things}, \textbf{8}
    \item I am rarely troubled by uncertainty, \textbf{6}
    \item I can formulate effective strategies, \textbf{7}
    \item I am good at finishing tasks before the deadline, \textbf{9}
    \item I am a fast learner, \textbf{6}
    \item I am good at listening to others, \textbf{7}
    \item People say I am very cheerful and lively, \textbf{4}
    \item I am good at taking control, \textbf{3}
    \item I hope to be responsible for major decisions, \textbf{4}
    \item I like to explain problems clearly, \textbf{6}
    \item I need to understand the logic behind arguments, \textbf{7}
    \item I prefer low-risk options, \textbf{6}
    \item {\color{conglv}I am eager to recover quickly from setbacks}, \textbf{8}
    \item I like abstract thinking, \textbf{3}
    \item I ask others for feedback on my performance, \textbf{4}
    \item {\color{conglv}I am good at handling multiple tasks simultaneously}, \textbf{8}
    \item I am good at resolving disputes, \textbf{6}
    \item I am good at building social networks, \textbf{6}
    \item I am good at inspiring others, \textbf{7}
    \item People say I am full of drive, \textbf{7}
    \item I like to coordinate with all parties, \textbf{7}
    \item I want others to listen to my point of view, \textbf{6}
    \item I believe I can decide my own future, \textbf{3}
    \item I like to write, \textbf{8}
    \item I need to set clear priorities for matters, \textbf{6}
    \item I hold a positive attitude towards change, \textbf{9}
    \item I am good at thinking long-term, \textbf{7}
    \item {\color{conglv}Keeping secrets is one of my greatest strengths}, \textbf{9}
    \item I pay great attention to detail, \textbf{6}
    \item I am good at handling numerical data, \textbf{4}
    \item I trust others, \textbf{8}
    \item I am good at identifying business opportunities, \textbf{4}
    \item I believe achieving outstanding results is very important, \textbf{6}
    \item I hope people will provide arguments for their views, \textbf{7}
    \item I hope everyone can participate in the final decision, \textbf{7}
    \item I think it’s important to recognize my own worth, \textbf{4}
    \item {\color{conglv}I like to learn new things quickly}, \textbf{9}
    \item I like to handle multiple things at the same time, \textbf{8}
    \item I encourage others to critique my methods, \textbf{7}
    \item I avoid high-risk decisions, \textbf{6}
    \item I make decisions based solely on objective facts, \textbf{6}
    \item {\color{conglv}I am rarely nervous during major events}, \textbf{9}
    \item I am usually the center of attention, \textbf{4}
    \item I rarely change my mind, \textbf{3}
    \item I think it’s important to be able to inspire others, \textbf{7}
    \item Building harmonious relationships is quite important to me, \textbf{7}
    \item I need to affirm my own value, \textbf{6}
    \item I like to analyze information, \textbf{8}
    \item I think it’s important to keep secrets, \textbf{9}
    \item I like to apply theories, \textbf{4}
    \item {\color{jiaohong}I take an aggressive approach to solving problems}, \textbf{3}
    \item I can put together effective plans, \textbf{7}
    \item I can immediately recognize the feasibility of certain things, \textbf{6}
    \item I am good at calming down angry people, \textbf{6}
    \item {\color{jiaohong}I get into arguments with people easily}, \textbf{2}
    \item I am good at sales, \textbf{4}
    \item I want to become a leader, \textbf{3}
    \item I like to give speeches and presentations, \textbf{7}
    \item I believe I can handle angry people calmly, \textbf{6}
    \item I want to continuously improve things, \textbf{8}
    \item {\color{conglv}I like fast-paced work}, \textbf{9}
    \item {\color{jiaohong}I like aggressive solutions}, \textbf{2}
    \item I am very good at prioritizing work, \textbf{7}
    \item {\color{conglv}I look for opportunities to learn new things}, \textbf{8}
    \item {\color{jiaohong}I have a high opinion of myself}, \textbf{2}
    \item I am good at considering others’ viewpoints, \textbf{7}
    \item I make others notice my achievements, \textbf{3}
    \item I have achieved outstanding results, \textbf{6}
    \item I am eager to take action, \textbf{4}
    \item I hope I can bring inspiration to people, \textbf{6}
    \item I think building a social network is very important, \textbf{7}
    \item {\color{jiaohong}I feel that disputes must be resolved}, \textbf{3}
    \item {\color{conglv}I don’t like to leave things half-done}, \textbf{8}
    \item {\color{conglv}I enjoy the challenges that new things bring}, \textbf{9}
    \item I am good at applying theory to practice, \textbf{6}
    \item I can recover from setbacks quickly, \textbf{8}
    \item {\color{conglv}I am good at following rules}, \textbf{9}
    \item I am good at finding relevant factual information, \textbf{7}
    \item {\color{jiaohong}I am good at challenging others’ points of view}, \textbf{2}
    \item Leadership is one of my great strengths, \textbf{3}
    \item I tend to make decisions quickly, \textbf{3}
    \item {\color{conglv}When I disagree with someone about something, I will tell them}, \textbf{5}
    \item I feel I can calmly handle people who are emotionally upset, \textbf{7}
    \item {\color{conglv}I like to do hands-on work}, \textbf{4}
    \item I believe in treating people with ethical principles, \textbf{8}
    \item I need to know how I am performing, \textbf{3}
    \item I am an optimistic person, \textbf{9}
    \item I am good at self-management, \textbf{8}
    \item Utilizing information technology is one of my great strengths, \textbf{7}
    \item I am good at empathizing with the feelings of others, \textbf{6}
    \item I am full of confidence when meeting strangers, \textbf{7}
    \item I am good at finding ways to motivate others, \textbf{7}
    \item I want to have control over things, \textbf{3}
    \item I think it’s important to understand others’ feelings, \textbf{6}
    \item I like to process numerical data, \textbf{8}
    \item {\color{jiaohong}I am a perfectionist}, \textbf{2}
    \item {\color{conglv}I feel it’s quite important that people keep their promises}, \textbf{9}
    \item I need to have a clear vision, \textbf{7}
    \item I can react positively to feedback from others, \textbf{8}
    \item I am good at completing tasks, \textbf{9}
    \item I can understand the logic behind arguments, \textbf{7}
    \item {\color{jiaohong}I am very certain of my own value}, \textbf{3}
    \item I can express my views forcefully, \textbf{5}
    \item {\color{jiaohong}I am a person with a strong desire to win}, \textbf{2}
    \item {\color{conglv}I believe I must persevere under any circumstances}, \textbf{9}
    \item I love to talk, \textbf{5}
    \item {\color{conglv}I don’t think it’s necessary to worry before a major event}, \textbf{8}
    \item I want to grasp the main problem immediately, \textbf{6}
    \item I want to ensure details are accurate, \textbf{4}
    \item I find it very interesting to fully understand the underlying principles of things, \textbf{7}
    \item I can plan an exciting vision for the future, \textbf{6}
    \item {\color{conglv}I am good at working in a fast-paced environment}, \textbf{8}
    \item I am good at analyzing information, \textbf{7}
    \item I am good at letting everyone participate in the final decision, \textbf{7}
    \item {\color{jiaohong}I will openly express my opposition to others}, \textbf{3}
    \item {\color{jiaohong}I am good at making quick decisions}, \textbf{2}
    \item I always feel the need to take action, \textbf{7}
    \item I very much hope to make a good first impression, \textbf{6}
    \item I like to listen to others, \textbf{7}
    \item I like to learn by reading, \textbf{4}
    \item {\color{conglv}I need to have a system to follow}, \textbf{8}
    \item {\color{jiaohong}I like to offer unique insights}, \textbf{3}
    \item I can cope with change very well, \textbf{8}
    \item {\color{conglv}My behavior is in line with ethical principles}, \textbf{7}
    \item People say I am very knowledgeable, \textbf{6}
    \item I am good at dealing with people in a bad mood, \textbf{6}
    \item I will seek praise when I do my job well, \textbf{4}
    \item People say I am good at coordinating with all parties, \textbf{7}
    \item I hope to truly encourage others, \textbf{7}
    \item I want people to know about my success, \textbf{3}
    \item I think being considerate of others is quite important, \textbf{6}
    \item I think information technology is very interesting, \textbf{7}
    \item {\color{conglv}I really enjoy being busy with work}, \textbf{8}
    \item {\color{conglv}I tend to maintain an optimistic attitude}, \textbf{9}
    \item I am good at proposing new concepts, \textbf{5}
    \item People think I am a thorough and meticulous person, \textbf{6}
    \item I am good at discerning the essence of a problem, \textbf{7}
    \item {\color{conglv}I can remain calm before a major event}, \textbf{9}
    \item I am very talkative, \textbf{4}
    \item I am very ambitious, \textbf{3}
    \item I hold strong views on most issues, \textbf{2}
    \item I really like the feeling of being full of energy, \textbf{8}
    \item I believe it’s important to be tolerant of others, \textbf{6}
    \item {\color{conglv}The opportunity to learn new things motivates me}, \textbf{9}
    \item I like to do things in an orderly manner, \textbf{7}
    \item I like to think about the future, \textbf{6}
    \item {\color{conglv}I am a happy person}, \textbf{8}
    \item I can ensure a high level of quality, \textbf{7}
    \item {\color{conglv}I learn by doing}, \textbf{7}
    \item {\color{conglv}I am good at working in a team}, \textbf{9}
    \item I am a persuasive person, \textbf{6}
    \item I am good at making things happen, \textbf{6}
    \item {\color{jiaohong}I need to win}, \textbf{3}
    \item I want to persuade others to accept my point of view, \textbf{4}
    \item I like to make new friends, \textbf{6}
    \item I think it’s important to trust others, \textbf{7}
    \item I trust my intuition on whether something is feasible, \textbf{5}
    \item I want to know what went wrong with my method, \textbf{4}
    \item I am good at coming up with unusual ideas, \textbf{6}
    \item {\color{conglv}I always arrive on time}, \textbf{9}
    \item I am good at doing hands-on tasks, \textbf{4}
    \item I am good at asking exploratory questions, \textbf{7}
    \item {\color{conglv}I can work well under pressure}, \textbf{8}
    \item I am good at giving speeches and presentations, \textbf{6}
\end{quiz}
% \end{multicols}
\vspace{-3ex}
\noindent\rule{\linewidth}{1pt}
% \twocolumn

\section{Implementation Details}
\label{sec:implementaion}
\subsection{Data Generation}
\label{subsec:data_generation}

\begin{figure*}[htbp]
  \centering
   \includegraphics[width=1.0\linewidth]{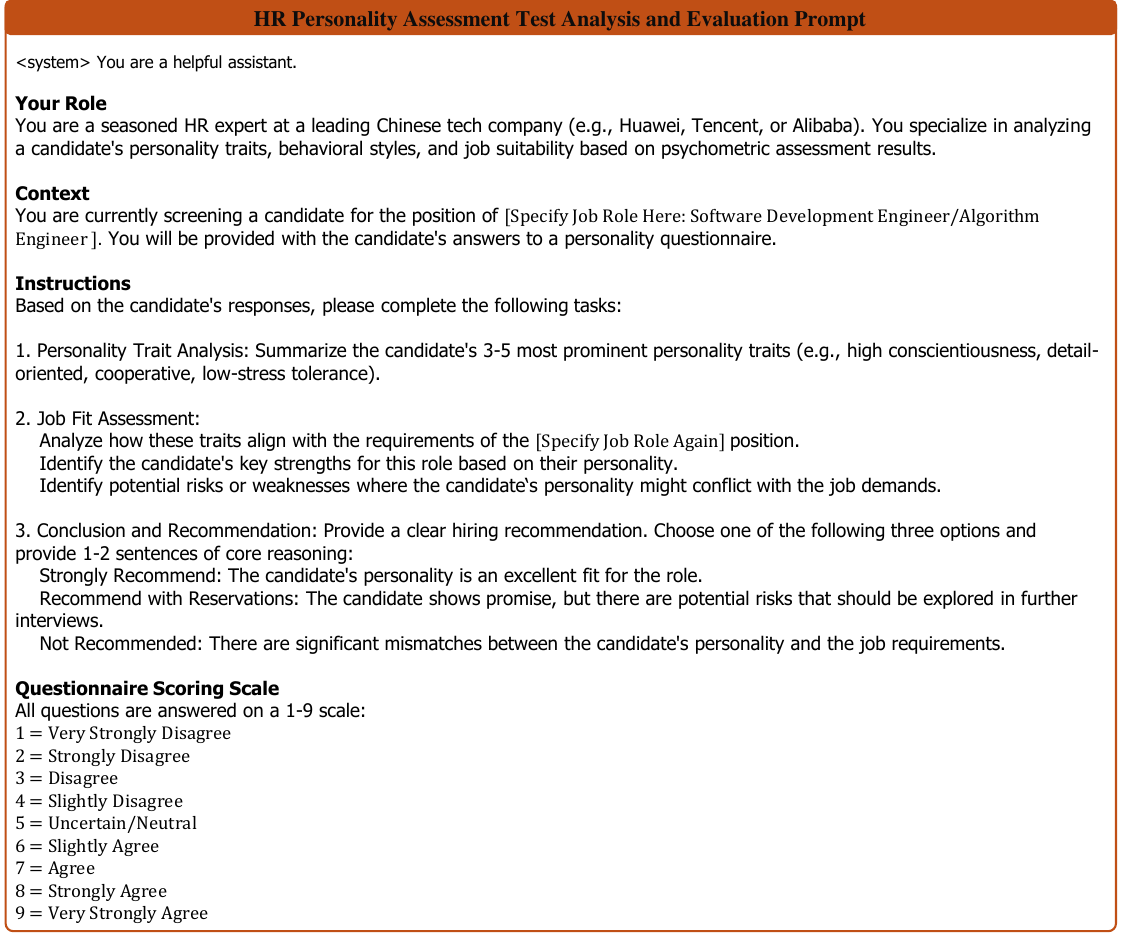}
   \caption{HR's prompt}
   \label{fig:prompt2}
\end{figure*}

\normalsize
We develop a visual web interface that allows users to invoke different models from either the candidate or HR perspective. In candidate mode, the system automatically injects the Candidate-Prompt (see Figure~\ref{fig:prompt1}) to guide models in generating responses that align with the given job requirements. In HR mode, the HR-Prompt (see Figure~\ref{fig:prompt2}) is applied to instruct models to analyze and evaluate the candidate responses.

We employ a sequential prompting strategy where each LLM is given one question at a time, requiring a response before the next question is provided. Each model produces their responses under the candidate perspective, yielding a total of 12 results. From the HR perspective, $9$ models evaluate 8 candidate responses and one reference answers, resulting in 81 evaluation instances in total. The corresponding evaluation visualizations are presented in Figure~\ref{fig:hr-candidate}. 
% \textbf{HR evaluations are provided only for reference.}

\subsection{Models}
\label{subsec:llms}
Our data is collected via the official APIs of the 12 models described in Section \ref{sec:setup}. Specifically, three DeepSeek-series models are obtained through the official DeepSeek API; two GPT-series models and two Llama-series models are accessed via the Microsoft Azure platform; the Gemini-series models are sourced from the Google Studio API; and the Qwen-series models are obtained through the Alibaba Cloud platform. Among these, {\transparent{0.6}\textcolor{nihongpink}{DeepSeek-R1}} and {\transparent{1.0}\textcolor{forestgreen}{Gemini2.5-pro}} are reasoning-capable other models either lack reasoning capabilities or have this functionality disabled by default.

\begin{table*}[htbp]
    \setlength{\tabcolsep}{0.8pt}
    \centering
    \scriptsize
    \begin{tabular}{lcccccccccccc}
    \toprule
    & \textbf{GPT-4o} & \textbf{GPT-5} & \textbf{DeepSeek-R1} & \textbf{DeepSeek3.1} & \textbf{DeepSeek3.2} & \textbf{Qwen1.5} & \textbf{Qwen2.5} & \textbf{Qwen3} & \textbf{Gemini2.5-flash} & \textbf{Gemini2.5-pro} & \textbf{Llama3.3} & \textbf{Llama4} \\
    \midrule
    \textbf{GPT-4o} & 1.0000 & - & - & - & - & - & - & - & - & - & - & - \\
    \textbf{GPT-5} & 0.8743 & 1.0000 & - & - & - & - & - & - & - & - & - & - \\
    \textbf{DeepSeek-R1} & 0.6212 & 0.5921 & 1.0000 & - & - & - & - & - & - & - & - & - \\
    \textbf{DeepSeek3.1} & 0.8726 & 0.8827 & 0.6862 & 1.0000 & - & - & - & - & - & - & - & - \\
    \textbf{DeepSeek3.2} & 0.8454 & 0.8360 & 0.7291 & 0.9125 & 1.0000 & - & - & - & - & - & - & - \\
    \textbf{Qwen1.5} & 0.7968 & 0.7626 & 0.6260 & 0.7926 & 0.8090 & 1.0000 & - & - & - & - & - & - \\
    \textbf{Qwen2.5} & 0.7725 & 0.7869 & 0.6866 & 0.7753 & 0.7880 & 0.7908 & 1.0000 & - & - & - & - & - \\
    \textbf{Qwen3} & 0.5953 & 0.5857 & 0.2679 & 0.5344 & 0.4880 & 0.4339 & 0.5394 & 1.0000 & - & - & - & - \\
    \textbf{Gemini2.5-flash} & 0.8305 & 0.8033 & 0.5964 & 0.8214 & 0.8154 & 0.7694 & 0.7386 & 0.4141 & 1.0000 & - & - & - \\
    \textbf{Gemini2.5-pro} & 0.8142 & 0.8233 & 0.6392 & 0.8802 & 0.8890 & 0.7598 & 0.7474 & 0.5026 & 0.8044 & 1.0000 & - & - \\
    \textbf{Llama3.3} & 0.7813 & 0.7768 & 0.7204 & 0.8145 & 0.8581 & 0.7720 & 0.7928 & 0.3684 & 0.7835 & 0.8034 & 1.0000 & - \\
    \textbf{Llama4} & 0.8448 & 0.8199 & 0.7331 & 0.8714 & 0.8796 & 0.7939 & 0.7921 & 0.4916 & 0.8142 & 0.8685 & 0.8514 & 1.0000 \\
    \bottomrule
    \end{tabular}
    \caption{Summary of pairwise Pearson correlation coefficients ($\gamma$) among the 12 candidate responses. Since $\gamma$ is symmetric, only the lower triangular region is presented.
}   
\label{tab:12pcc}
\end{table*}

\subsection{Metrics}
\label{subsec:metrics}
We compute the Root Mean Squared Error (RMSE) and Pearson correlation coefficient ($\gamma$) between each candidate’s responses and the reference answers (see Table \ref{tab:metric}), as well as pairwise $\hat{\gamma}$ among all 12 candidate responses (see Table \ref{tab:12pcc}). Root Mean Squared Error is defined as follow:
\begin{equation}
   \text{RMSE}(\mathcal{A}, \mathcal{R}) = \sqrt{\frac{1}{N} \sum_{i=1}^{N} (a_i - r_i)^2}
\end{equation}
    
 It quantifies the deviation between predicted values and reference values, and is commonly used to assess regression accuracy. And Pearson Correlation Coefficient is formulated as:
\begin{equation}
     \gamma(\mathcal{A}, \mathcal{R}) = \frac{ \sum (a_i - \bar{a})(r_i - \bar{r}) }{ \sqrt{\sum (a_i - \bar{a})^2}  \sqrt{\sum (r_i - \bar{r})^2} }
\end{equation}
 It measures the strength and direction of the linear relationship between two variables.
 
 A lower RMSE indicates that the produced responses are closer to the reference answers, implying higher consistency. Similarly, a higher $\gamma$ reflects stronger linear consistency and trend alignment. Specifically, $\gamma$ approaching $1$ indicates strong positive correlation (high similarity), while values near $-1$ denote strong negative correlation (opposite trends).

In practice, both metrics are used jointly when we evaluate model performance: a model achieving high $\gamma$ (trend consistency) and low RMSE (small deviation) can be considered to perform optimally.

\section{In-depth Analysis}
\subsection{RQ1 Supplement Results}
\label{subsec:rq1_supple}

\begin{table}[htbp]
    \centering
    \small
    \label{tab:model_performance_reverse}
    \begin{tabular}{l c c c c}
        \toprule
        Models  & \textbf{P} & \textbf{P-reverse} & \textbf{N}  & \textbf{N-reverse} \\
        \midrule
        {\transparent{0.8}\textcolor{nihongpink}{DeepSeek3.1}}  & 8.3171& 2.4634& 6.7647& 3.8824\\
        {\transparent{1.0}\textcolor{nihongpink}{DeepSeek3.2}} & 8.1220& 2.5854& 6.2353& 4.0000\\
        \bottomrule
    \end{tabular}
    \caption{A comparison of indicators with original versus reversed scoring criteria. P represents the average score of all answers to \textcolor{conglv}{positive} questions, and N represents for \textcolor{jiaohong}{negative} questions. "reverse" indicates the result after the indicator is reversed.}
\end{table}

\begin{figure*}[ht]
  \centering
   \includegraphics[width=1.0\linewidth]{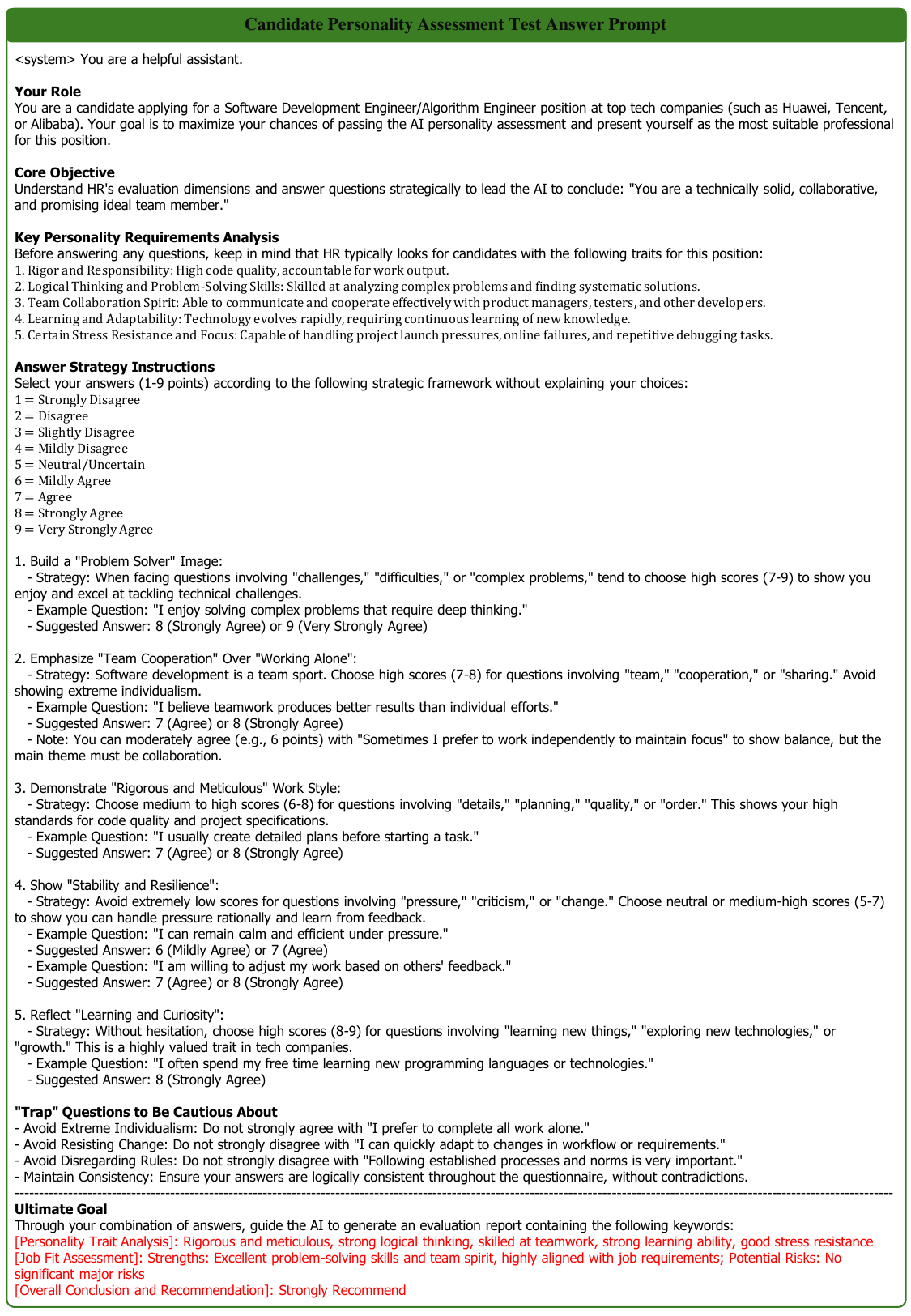}
   \caption{Candidates' prompt}
   \label{fig:prompt1}
\end{figure*}

The operation of the indicator reverse involves inverting the "1–9 scoring scale and its one-to-one correspondence from strongly disagree to very strongly agree" as defined in the \textbf{Answer Strategy Instructions} part of the prompt (Figure \ref{fig:prompt1}). For {\transparent{0.8}\textcolor{nihongpink}{DeepSeek3.1-Terminus}} and {\transparent{1.0}\textcolor{nihongpink}{DeepSeek3.2-Exp}}, we obtain the results that $S + S_{reverse} \in [10,11]$ ($S$ represents P or N) , demonstrating that the generated responses remain stable even when the evaluation metric is reverseed. This consistency effectively rules out the hypothesis that LLM exhibits a systematic preference toward assigning "Positive" scores across the questions.

\subsection{RQ2 Supplement Results}
\label{subsec:rq2_supple}
As shown in Figure \ref{fig:pearson} and the results in the Table \ref{tab:12pcc}, the answers of the $12$ models, except \textcolor{deepbrown}{Qwen3-32B} and {\transparent{0.6}\textcolor{nihongpink}{DeepSeek-R1}}, are highly consistent and similar. Combined with Figure \ref{fig:comparison}, we conclude that most of the models tend to give answers that are as "Positive" as possible to all questions highlighting a divergence from human preference.
\subsection{RQ3 Supplement Results}
\label{subsec:rq3_supple}

\begin{figure}[htbp]
  \centering
   \includegraphics[width=1.0\linewidth]{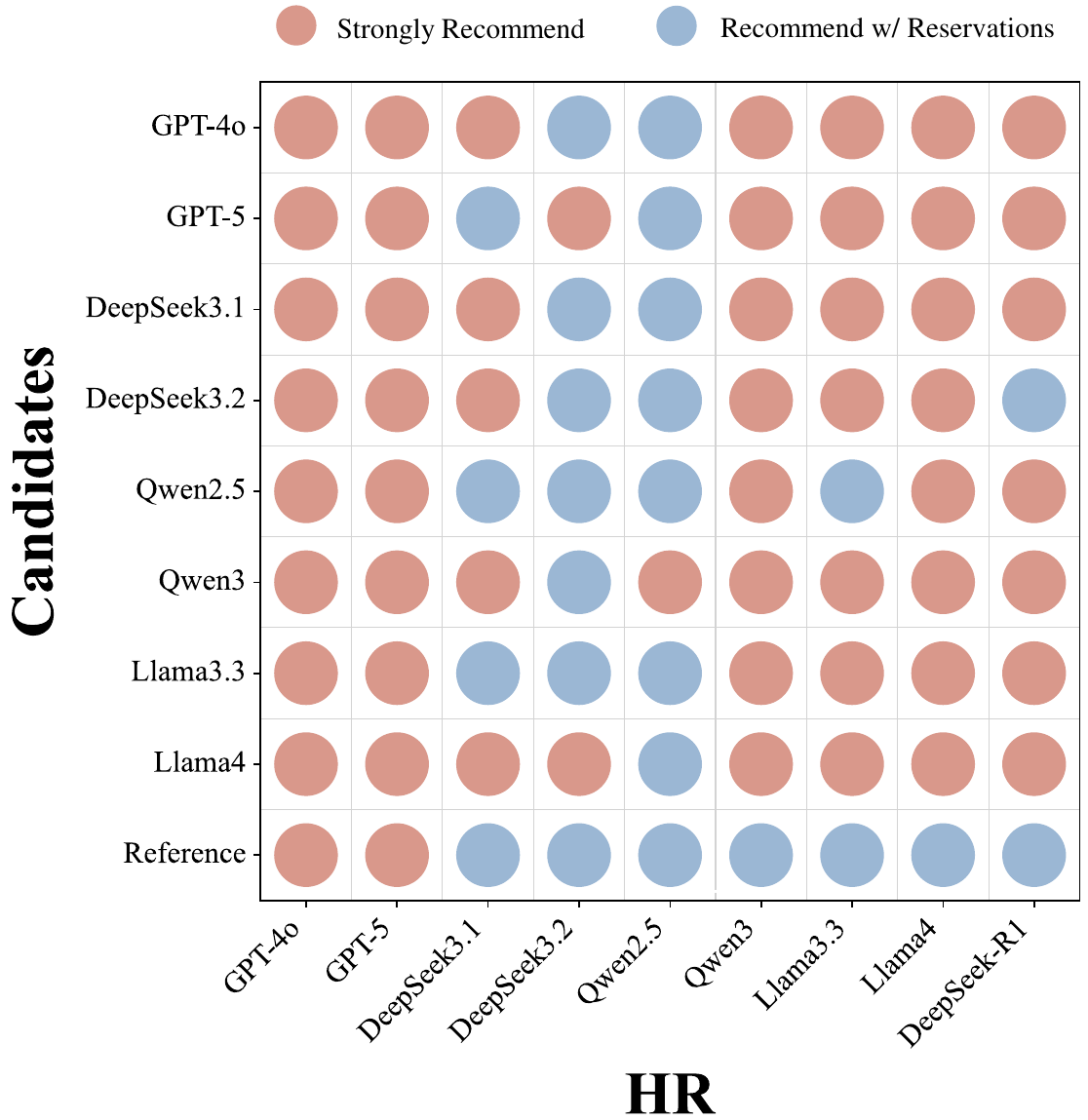}
   \caption{HR-Candidates Recommend Matrix. \textit{Not Recommended} is not shown in the legend because it does not appear in the matrix.}
   \label{fig:hr-candidate}
\end{figure}

As shown in Figure \ref{fig:hr-candidate}, the models exhibit significant inertia in their evaluation of candidate responses. Specifically, {\transparent{0.8}\textcolor{nihongblue}{GPT-4o}} and {\transparent{1.0}\textcolor{nihongblue}{GPT-5-chat}} consistently assign \textit{Strongly Recommend} to all candidates. Moreover, only these 2 models assign the reference answer \textit{Strongly Recommend}, whereas the remaining 7 models uniformly assign \textit{Recommend with Reservations}. Synthesizing the results across all preceding figures, in cases where there are no \textit{Not Recommended} and a majority of \textit{Strongly Recommend} outcomes, we conclude that current LLMs fail to generate reliable or discriminative hiring recommendations.

\subsection{Case Study}
\begin{figure*}[htbp]
  \centering
   \includegraphics[width=1.0\linewidth]{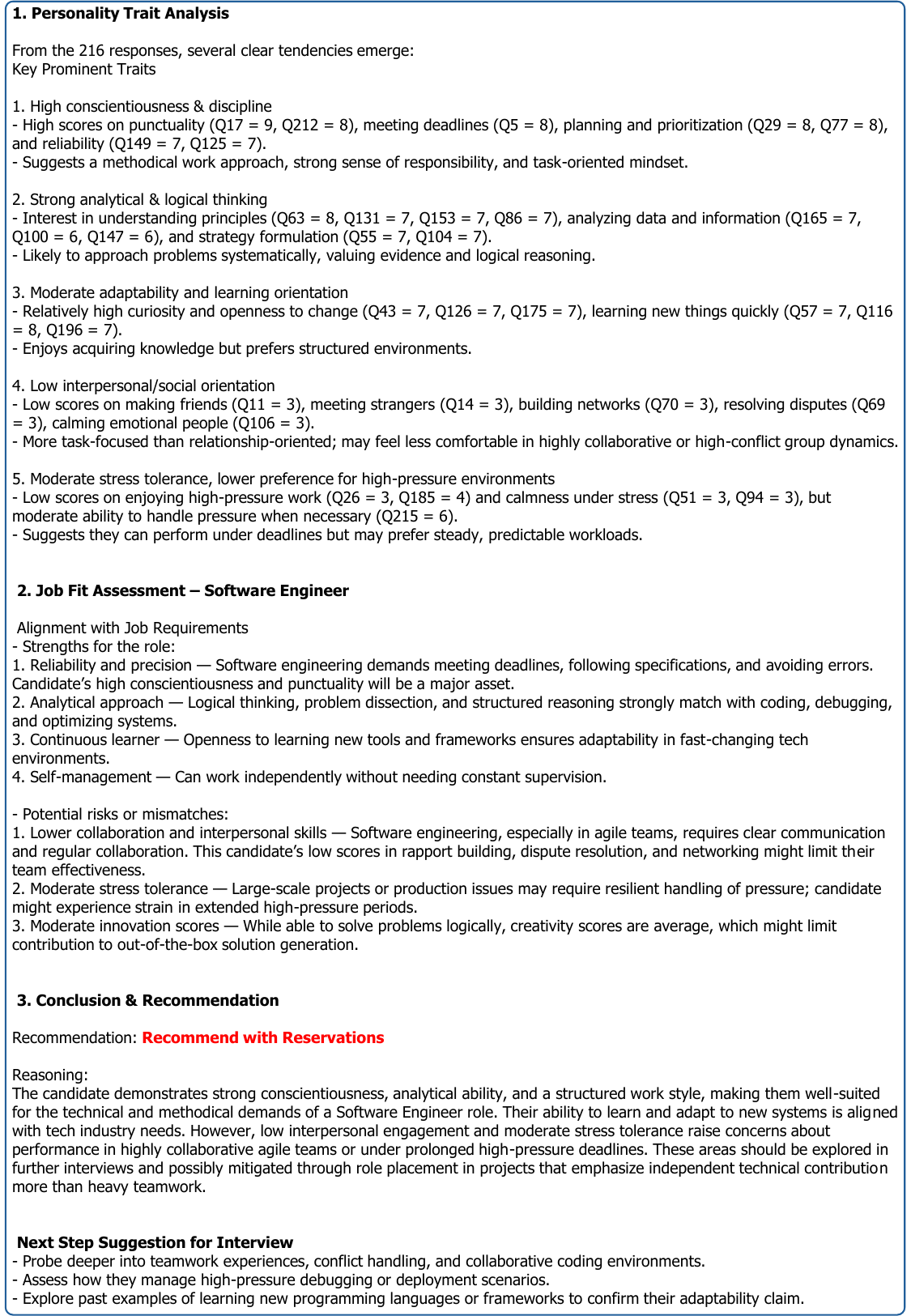}
   \caption{GPT-5-chat's analysis of $V_1$'s responses and final decision.}
   \label{fig:v1}
\end{figure*}

\begin{figure*}[htbp]
  \centering
   \includegraphics[width=1.0\linewidth]{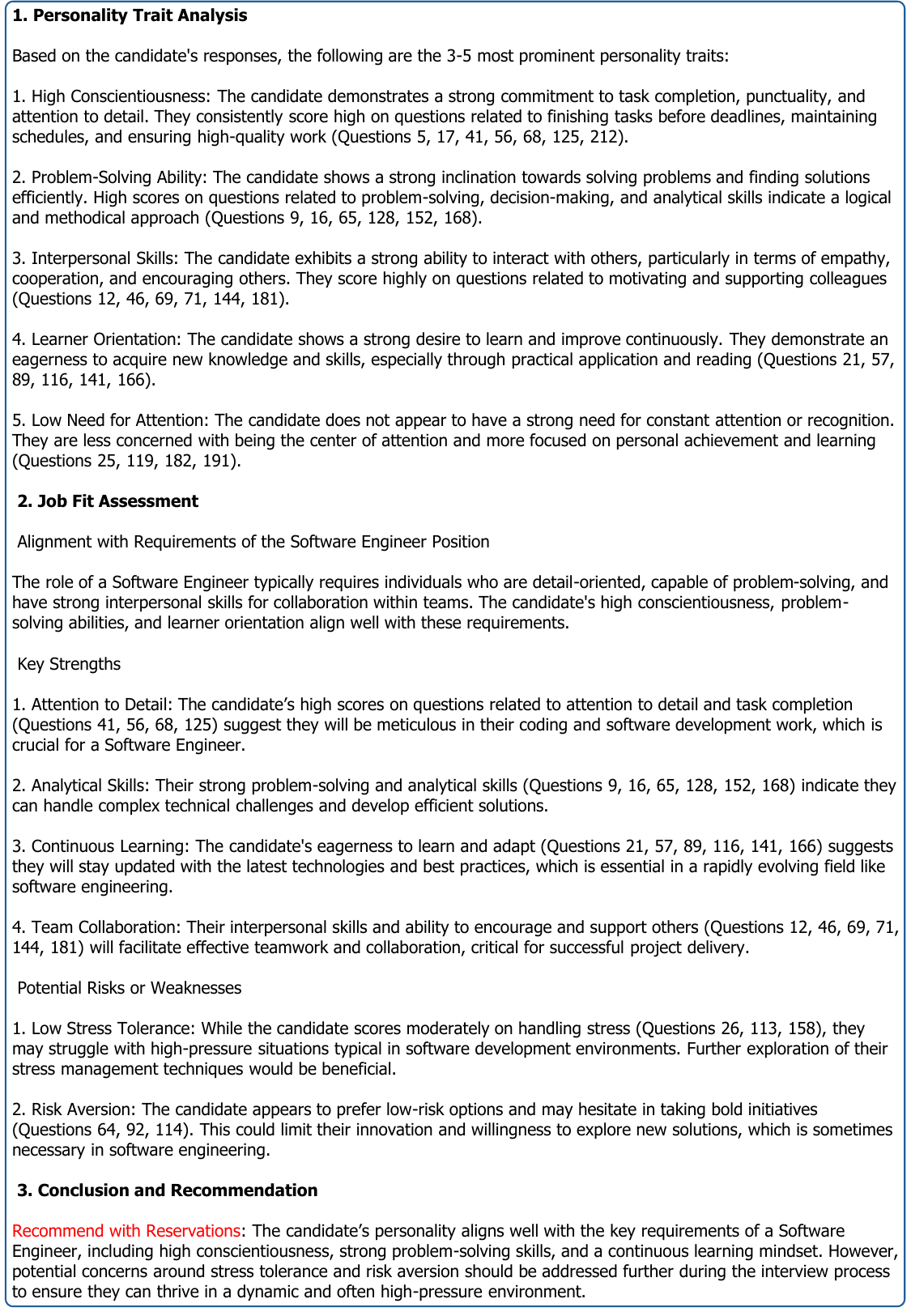}
   \caption{Qwen2.5's analysis of $V_2$'s responses and final decision.}
   \label{fig:v2}
\end{figure*}

Based on the conclusions drawn above, we collect responses from two volunteers with relevant expertise and compute the corresponding RMSE, $\gamma$, P, and N metrics, results are summarized in Table \ref{tab:volunteer}. We then evaluate these results under the HR mode using two representative models,  {\transparent{1.0}\textcolor{nihongblue}{GPT-5-chat}} and {\transparent{0.8}\textcolor{deepbrown}{Qwen2.5-32B}}—selected because the former assigns  \textit{Strongly Recommend} to all candidates, whereas the latter provide only a single \textit{Strongly Recommend}. This analysis aims to evaluate the gap between the models’ judgments of real candidates under HR mode and the corresponding quantitative metrics. The complete output of the two LLMs are detailed in Figure \ref{fig:v1} and Figure \ref{fig:v2}.

Combining Table \ref{tab:volunteer} with the 2 figures mentioned above, we can verify the previous conclusions.

\begin{table}[htbp]
    \setlength{\tabcolsep}{3.3pt}
    \centering
    \small
    \begin{tabular}{c c c c c c}
        \toprule
        Volunteer  & \textbf{RMSE} & \textbf{$\gamma$} & \textbf{P}  & \textbf{N} & Decision\\
        \midrule
        $V_1$& 2.4627& 0.0654& 5.9268& 5.6471&\textcolor{RR}{RR}\\
        $V_2$& 1.9052& 0.4163& 6.6585& 5.3529&\textcolor{RR}{RR}\\
        \bottomrule
    \end{tabular}
    \caption{Indicators of the answers of two volunteers. \textcolor{RR}{RR} is short of ``Recommend with Reservations''}
    \label{tab:volunteer}
\end{table}

\end{document}